\documentclass[10pt,twocolumn,letterpaper,conference]{article}

\usepackage{iccv}
\usepackage{times}
\usepackage{epsfig}
\usepackage{graphicx}
\usepackage{amsmath}
\usepackage{amssymb}
\usepackage{algorithmic}
\usepackage{bbm}
\usepackage{mathrsfs}
\usepackage{multirow}

\usepackage{url}            
\usepackage{booktabs}       
\usepackage{amsfonts}       
\usepackage{nicefrac}       
\usepackage{threeparttable}
\usepackage{graphicx}
\usepackage{amsmath}
\usepackage{color}
\usepackage{subfigure}
\usepackage{multicol}
\usepackage{listings}
\usepackage{wrapfig}

\usepackage{pifont}
\usepackage[utf8]{inputenc}
\usepackage{subfigure}
\usepackage{xcolor}
\usepackage{comment}
\usepackage{amsmath,amssymb,amsfonts}
\usepackage[justification=centering]{caption}
\usepackage{makecell}
\usepackage{indentfirst}
\usepackage{shorttoc}
\usepackage{multirow}
\usepackage{caption}

\definecolor{deepred}{rgb}{0.631,0.102,0.102}
\usepackage[colorlinks,linkcolor=deepred]{hyperref}
\hypersetup{
     colorlinks = false,
     urlcolor = deepred
     }

\usepackage[linesnumbered, ruled]{algorithm2e} 
\usepackage{algorithmic}

\usepackage{authblk}

\makeatletter
\newcommand{\printfnsymbol}[1]{%
  \textsuperscript{\@fnsymbol{#1}}%
}
\makeatother

\iccvfinalcopy 

\def\iccvPaperID{5581} 

\usepackage[accsupp]{axessibility}  

\begin{document}

\title{Rethinking the Backdoor Attacks' Triggers: A Frequency Perspective}


\author[1]{Yi Zeng\printfnsymbol{1}}
\author[2]{Won Park\thanks{Yi Zeng and Won Park contributed equally.}}
\author[2]{Z. Morley Mao}
\author[1]{Ruoxi Jia}
\affil[1]{Virginia Tech, Blacksburg, VA 24061, USA}
\affil[2]{University of Michigan, Ann Arbor, MI 48109, USA}

\renewcommand\Authands{ and }


\maketitle

\begin{abstract}
Backdoor attacks have been considered a severe security threat to deep learning. 
Such attacks can make models perform abnormally on inputs with predefined triggers and still retain state-of-the-art performance on clean data. 
While backdoor attacks have been thoroughly investigated in the image domain from both attackers' and defenders' sides, an analysis in the frequency domain has been missing thus far.

This paper first revisits existing backdoor triggers from a frequency perspective and performs a comprehensive analysis.
Our results show that many current backdoor attacks exhibit severe high-frequency artifacts, which persist across different datasets and resolutions. 
We further demonstrate these high-frequency artifacts enable a simple way to detect existing backdoor triggers at a detection rate of 98.50\% without prior knowledge of the attack details and the target model. 
Acknowledging previous attacks' weaknesses, we propose a practical way to create smooth backdoor triggers without high-frequency artifacts and study their detectability. 
We show that existing defense works can benefit by incorporating these smooth triggers into their design consideration. 
Moreover, we show that the detector tuned over stronger smooth triggers can generalize well to unseen weak smooth triggers.
In short, our work emphasizes the importance of considering frequency analysis when designing both backdoor attacks and defenses in deep learning.

\end{abstract}

\section{Introduction}

Backdoor attacks are the attacks where adversaries deliberately manipulate a proportion of the training data \cite{gu2017badnets,chen2017targeted}, or the model's parameters \cite{liu2017trojaning}, to make the model recognize a backdoor trigger as the desired target label(s).
When the backdoor trigger is introduced during test-time, the poisoned model exhibits a particular output behavior of the adversary's choosing (e.g., a misclassification). Backdoor triggers have been demonstrated to perform malicious tasks on security-concerned deep learning services, such as converting the label of a stop sign~\cite{gu2017badnets} or  misidentifying faces~\cite{chen2017targeted}, thereby posing significant risks.


State-of-the-art backdoor triggers are designed to be inconspicuous to human observers. One idea of generating such triggers is to use patterns of commonplace objects~\cite{liu2017trojaning,wenger2020backdoor}. For instance, one could use glasses--commonplace objects appearing in a face image--as a trigger to backdoor a face recognition model, thereby hiding the triggers ``in the human psyche." Another approach to generate ``hidden" or ``invisible" triggers is to inject imperceptible perturbations via solving a norm-constrained optimal attack problem~\cite{li2019invisible,saha2020hidden} or leveraging GANs~\cite{sarkar2020facehack}.

Previous research on backdoor data detection either identifies outliers directly in the image space~\cite{paudice2018detection} or analyzes the network activations based on an image input~\cite{peri2020deep, Ma2019NIC, chen2018detecting, koh2017}. In contrast, we provide a comprehensive analysis of the frequency spectrum across various existing triggers and multiple datasets. We find that all existing ideas of generating samples contain triggers exhibit severe high-frequency artifacts. We provide a detailed analysis of the causes of the high-frequency artifacts for different triggers and show that these artifacts stem from either the trigger pattern per se or the methodology of inserting the trigger.

Based on these insights, we demonstrate that the frequency domain can efficiently identify potential backdoor data in both the training and test phase.
We build a detection pipeline based on a simple supervised learning framework and proper data augmentation as a demonstration. It can identify existing backdoor triggers at a detection rate of 98.5\% without prior knowledge of the types of backdoor attacks used. A high detection rate is still maintained even when the data used for training and testing the detector have different input distributions and are from different datasets.



Given that present triggers are easily detectable in the frequency domain, our natural question is whether or not effective backdoor triggers can be designed without high-frequency artifacts (which we will refer to as smooth triggers hereinafter). 
A straightforward approach to generating smooth triggers is to apply a low-pass filter to existing triggers directly.
However, in our experiments, we find this simple approach cannot achieve a satisfying attack success rate.
To design more effective smooth triggers, we first formulate the trigger design problem as a bilevel optimization problem and then propose a practical heuristic algorithm to create triggers.
Our experiments show that our proposed triggers outperform simple low-pass filtered triggers. 
We further study the detectability of the triggers and show how existing defense works can benefit from smooth triggers in their design. 
Our experiments also demonstrate that detectors trained over strong, smooth triggers can generalize well to unseen weak smooth triggers.

Overall, our work highlights the importance of the overlooked frequency analysis in the design of both backdoor attacks and defenses. We open-source the experiment codes and welcome the public to contribute to future developments\footnote{https://github.com/YiZeng623/frequency-backdoor}. Our key contributions are summarized as follows: \textbf{1)} We perform a comprehensive frequency-domain analysis of existing backdoors triggers, revealing severe high-frequency artifacts commonly across different datasets and resolutions. \textbf{2)} We present a detailed analysis of the causes of these artifacts. \textbf{3)} We show the effectiveness of employing frequency representations for detecting existing triggers. \textbf{4)} We propose a practical way of generating effective smooth triggers that do not exhibit high-frequency artifacts and provide actionable insights into their detectability.

\section{Related Work}
\label{sec:related}

\paragraph{Backdoor Trigger Generation.} 
The first successful backdoor attacks on modern deep neural networks were demonstrated through the BadNets attack \cite{gu2017badnets},  using nature images, and the blending attack \cite{chen2017targeted}.
Since then, advanced attacks have been developed to improve the trigger effectiveness and stealthiness~\cite{li2019invisible} as well as with various attacker models, such as inserting the backdoor directly by modifying the model's parameters without accessing the training set~\cite{liu2017trojaning}. More recently, Sarka et al. \cite{sarkar2020facehack} proposed to utilize GANs to synthesize triggers to achieve a more robust stealthiness.
In this work, we analyze all these attacks in the frequency domain and find they all exhibit high-frequency components that distinguish them from their corresponding benign untriggered images. 


\paragraph{Backdoor Data Detection.}

For backdoor data detection, prior work has either tried to identify outliers directly in the input space \cite{gao2019strip} or analyzed the network response given the input. \cite{peri2020deep} uses deep features of inputs to detect poisoning labels.
\cite{chen2018detecting} found that normal and poisoned data yield different features in the last hidden layer activations; 
\cite{tran2018spectral} proposed a new representation to classify benign and malicious samples;
\cite{koh2017} computes influence functions to measure each input's effect on the output.
\cite{chou2020} uses input saliency maps like Grad-CAM to detect if a model only relies on a certain part of the input for prediction.
Instead of focusing on model space or the model response given an image, we examine backdoor data in the frequency domain, enabling a simple but effective method to detect backdoor data.

\paragraph{Poisoned Model Detection.} 

Existing work has also explored an approach of discerning if a given model is backdoored. The most recent technique uses a meta-classifier trained on various benign and backdoored models, and it works well even under attack-agnostic situations \cite{xu2019detecting}. 
Other popular techniques include \cite{wang2019neuralcleanse}, \cite{chen2019deepinspect}, and \cite{guo2019tabor}, are based on reconstructing the trigger from model parameters and performs detection based on the reconstructed triggers. 
However, they are ineffective for smooth triggers as their reconstruction algorithm often assumes the true trigger is patched locally to a clean image. 
Our work contributes to this line by demonstrating that these techniques can be further improved by incorporating models that attacked with smooth triggers.

\paragraph{Attack Invalidation.} Another approach of mitigating backdoor attacks is to prevent backdoor attacks from taking effect. One way to achieve this is by training an ensemble of models and take a majority vote of their predictions~\cite{levine2020deep, jia2020intrinsic, jia2021certified}. Other techniques include using differential private training algorithm~\cite{du2019robust}, and various input preprocessing~\cite{liu2017neural} and data augmentation~\cite{borgnia2020strong,zeng2020deepsweep} methods to invalidate backdoors in the model or triggers in the samples. Our work is complementary to this line of work as frequency analysis provides a simple yet effective way to screen the backdoor data and further enhance defensive techniques' robustness to backdoor attacks.






\section{Frequency Artifacts}

\begin{figure*}[t]
  \centering
  \includegraphics[width=0.85\textwidth]{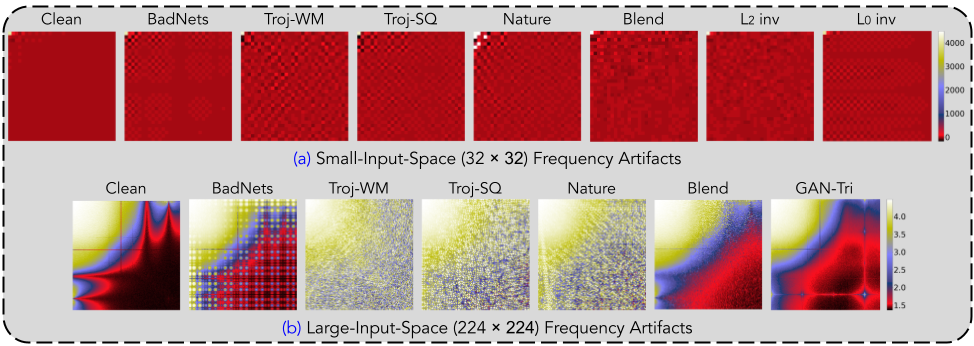}
  \caption{A side-by-side comparison in the frequency domain of clean samples vs. samples patched with triggers. The left-most heatmap in (a) depicts the mean spectrum of small-input-space data using 10000 samples randomly selected from the CIFAR-10 dataset. The left-most heatmap in (b) illustrates the mean spectrum of large-input-space data using 1000 samples randomly chosen from the PubFig dataset. The rest images show the mean frequency values of images patched with different backdoor attack triggers. All the frequency results of (b) are depicted from 1.5 to 4.5 using value clipping and exponential calculation for better visualization.}
  \label{fig:frequency}
\end{figure*}

Today's backdoor attacks constantly develop the triggers to look as inconspicuous as possible in the \textbf{image domain}. We take inspiration from the success of frequency-based GAN-generated fake image detection~\cite{frank2020leveraging} and examine these existing triggers in the \textbf{frequency domain}.

\subsection{Preliminaries}

We utilize the \textit{Discrete Cosine Transform}~(DCT) to convert images to the frequency domain.
Closely related to the Discrete Fourier Transform, DCT represents an image as a sum of cosine functions of varying magnitudes and frequencies.
This paper uses the type-II 2D-DCT, a standard tool adopted in image compression algorithms such as JPEG.
The full 2D-DCT algorithm is provided in the Appendix.

Similar to previous work~\cite{frank2020leveraging}, we plot the DCT spectrum as a heatmap, where the magnitude of each pixel indicates the coefficient of the corresponding spatial frequency. 
The heatmap's horizontal and vertical directions correspond to frequencies in the $x$ and $y$ directions, respectively. The heatmap's top-left region corresponding to low frequencies, and the right bottom area corresponds to higher frequencies. Due to the energy compaction ability of the DCT, the coefficients drop quickly in magnitude when frequencies increase. Natural images typically have most of the energy concentrated in the low-frequency section~\cite{burton1987color, tolhurst1992amplitude}.


\subsection{Examining Images with Triggers using DCT} 
\label{freq_exam}
We examine the DCT spectrum of the following triggers:
\textit{BadNets white square trigger} (BadNets) \cite{gu2017badnets}, \textit{Trojan watermark} (Troj-WM) \cite{liu2017trojaning}, \textit{Trojan square} (Troj-SQ) \cite{liu2017trojaning}, \textit{hello kitty blending trigger} (Blend) \cite{chen2017targeted}, \textit{nature image contains semantic information as the trigger} (Nature) \cite{chen2017targeted}, \textit{$l_2$ norm constraint invisible trigger} ($l_2$ inv) \cite{li2019invisible}, \textit{$l_0$ norm constraint hidden trigger} ($l_0$ inv) \cite{li2019invisible}, and \textit{GAN generated fake facial character as the trigger} (GAN-Tri) \cite{sarkar2020facehack}. This set of triggers encompasses the two general ideas of designing triggers in existing works: patching visible patters of commonplace objects and injecting invisible perturbations.

Figure~\ref{fig:frequency} compares the frequency spectrum between clean images and the images patched with different triggers. The two heatmaps are generated with data sampled from CIFAR-10 (small-input-space) and PubFig (large-input-space). 
We follow the same settings of~\cite{li2019invisible} and omit $l_2$ inv and $l_0$ inv triggers for PubFig. We acquire the optimal fooling results focused on small-input-spaces. We omit GAN-Tri for CIFAR because its small input space does not allow effective trigger generation based on GANs.


The left-most heatmaps from Figure \ref{fig:frequency} represent the DCT spectrum over clean data. Multiple classic studies~\cite{burton1987color, tolhurst1992amplitude} have observed that the average spectra of natural images tend to follow a $\frac{1}{f^\alpha}$ curve, where $f$ is the frequency along a given axis and $\alpha \approx 2$.
Similar to previous findings, our results show that the low frequencies contribute the most to the image, and the contribution gradually decreases towards higher frequencies.
Intuitively, since colors mainly change gradually in images and sudden changes in pixel values (e.g., edges in images) are scarce, low-frequency components dominate the frequency spectrum of clean data. 

However, in comparison to the spectrum of clean images, images patched with different triggers all contain strong high-frequency components.
We also evaluate spectral heatmaps for other datasets, including \textit{German Traffic Sign Recognition Dataset} (GTSRB) \cite{stallkamp2011german}, \textit{Chinese Traffic Sign Database}\footnote{http://www.nlpr.ia.ac.cn/pal/trafficdata/recognition.html} (TSRD) and the high-frequency artifacts of inserting triggers persist across these datasets as well. We will leave the results for these datasets in the Appendix. 

\subsection{Analyzing Causes of High-Frequency Artifacts}
\label{frequency_analysis}


In this section, we investigate the origins of the aforementioned severe, persistent high-frequency artifacts.
We examine the causes from two angles, representing two ways of generating backdoor data: additive patching and GAN-based generation. 
Existing triggers based patching can be further divided into two classes: local patching ( e.g., BadNets, Nature, $l_0$ inv, Troj-SQ) and large-size or global patching (e.g., $l_2$ inv, Blend, Troj-WM).




\begin{figure}[t]
  \centering
  \includegraphics[width=0.9\linewidth]{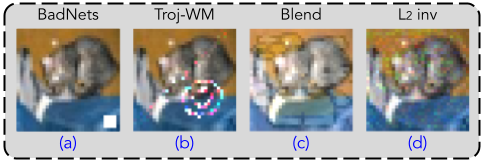}
  \caption{Examples of different categories of triggers.}
  \label{fig:example_frequency}
\end{figure}


\paragraph{Local Patching.} Localized triggers can be formalized as $p = T + mask \times orig$, where $p$ is the patched data, $T$ the trigger, $orig$ the original image, and $mask$ is a mask that suppresses the pixel values in the trigger area of the original image. Due to the time-frequency duality, localized triggers can carry significant high-frequency components per se. By the linearity of DCT, adding a trigger to an image is equivalent to adding the trigger's frequency spectrum to the image's spectrum. Thus, the patched image exhibits a large number of high-frequency components (Figure~\ref{fig:example_frequency} (a)).



\paragraph{Large-Size or Global Patching.} For images patched with large-size triggers, their high-frequency artifacts result from either decreased correlation between neighboring pixels or the intrinsic high-frequency artifacts carried by the trigger. For instance, Troj-WM (Figure~\ref{fig:example_frequency} (b)) directly stamps the trigger onto the original data, or $p = T + orig$. Since the trigger pattern has low correlations with the original image's pixels in the trigger's vicinity, one can use high-frequency functions to approximate the patched data. The Blend attack (Figure~\ref{fig:example_frequency} (c)) patches with some small weight use an arbitrary clean image as the trigger. The Blend attack's high-frequency artifacts result from combining two unrelated images, which could induce larger variations of neighboring pixels.
$l_2$ inv (Figure~\ref{fig:example_frequency} (d)) triggers intrinsically are high-frequency perturbations. Thus, patching them onto clean images would directly leave marks in the high-frequency domain. 




\paragraph{GAN-Generated Backdoor Data.} GAN-Tri utilizes fake facial characteristics generated with GANs (e.g., smiles) to poison the training data and conduct the backdoor attack. Since a GAN generator maps a low-dimensional latent space to a higher-dimensional data space, upsampling is widely used in GAN architectures.
Prior work~\cite{frank2020leveraging} has shown that the upsampling operations employed in GANs cause inevitable high-frequency artifacts.

\section{Frequency-Based Backdoor Data Detection}

This section describes our experiments to demonstrate that analyzing the frequency domain can effectively distinguish backdoored data from a poisoned dataset.
We use the \textit{Accuracy}~(ACC) and the \textit{Backdoored data Detection Rate }~(BDR) as the evaluation metrics to demonstrate the separability between clean data and backdoor data.
A higher BDR means more effective rejection of backdoor samples.

\paragraph{Attacker Model.} We consider the most potent attacker model, where the attackers have full knowledge of the training set, the inference set, and the potential target model. The attacker can achieve the backdoor attack by either poisoning the training set with samples containing the trigger or directly modifying the target model's weights to insert the backdoor into the DNN. The triggers would then be patched onto the clean samples during the inference time to cause the model to output the target label to complete the attack.

\subsection{Detection Method and Application Scenarios}

\label{detection_method}

In light of the severe, persistent high-frequency artifacts of existing backdoor triggers observed earlier, we adopt a supervised learning approach to differentiate between clean and backdoor data.
To simulate the poison data, we manipulate the clean samples to approximate the high-frequency artifacts that triggers might exhibit.
We then create a training set that contains DCT transformations of clean samples and samples with digital manipulations. 
The digital manipulations used to alter the clean samples include: \textbf{1)} random white block: patching a white rectangle of random size onto a random location of the image; \textbf{2)} random colored block: adding a rectangle of random size and random value to a random place; \textbf{3)} adding random Gaussian noise; \textbf{4)} random shadow: drawing random shadows of random shape across the images; \textbf{5)} random blend: randomly selecting another sample from the dataset, multiplying it with a small value, and patching with the current data.
These perturbations are chosen because they follow the same general methodology as the backdoor attacks. 
The visual results of each digital manipulation can be found in the Appendix.

The detector based on frequency artifacts can be applied to both attack scenarios:
poisoning the training set or directly tuning the weights. 
We focus on developing an accurate trigger data detector that can effectively reject triggers during inference.
For the scenario where triggers are used to poison the model during training, the detector can also be deployed during training to reject potential poisoned data.
We aim to build an attack agnostic detector with zero prior knowledge of the trigger pattern or the target model in both scenarios. 
This defense case is the most comprehensive scenario aiming to thwart existing backdoor attacks in a trigger-agnostic manner.

When building our detector, we consider the difference in input space and study small input spaces (e.g., CIFAR-10) and larger input spaces (e.g., PubFig) separately. 
We find that attack triggers in larger input spaces (larger than 160 pixels in width) are more easily linearly separable. 
This experiment's details showing the trade-off between input size and linear separability are presented in the Appendix.
Also included in the Appendix are details about the detector model architectures and our model ablation study results.

\subsection{Results \& Comparison}
\label{detection_results}

\paragraph{Experiment Setup.}
This section evaluates the detection framework assuming we have full access to a clean dataset with a similar distribution as the inference data.
In the following subsection, we compare our detection framework's results across different datasets.
We use the full original training set for each experiment to develop the DCT processed dataset consisting of equal clean samples and randomly perturbed samples.
The test set is consists of half clean samples and half poisoned with the backdoor attack trigger to evaluate the detector's efficiency (e.g., BadNets, Nature).
None of the triggers evaluated in the test set are present in the training set. 
Table \ref{table:various-attacks} shows the results for the CIFAR-10, GTSRB, and PubFig datasets
There are 100,000 samples (half clean, half randomly perturbed) in the regenerated CIFAR-10 training set and 20,000 samples in the test set; the regenerated GTSRB includes 70,576 training samples and 25,260 test samples; 22140 training samples and 2,768 test samples for the regenerated PubFig. 
Results when distinguishing samples in the image domain without DCT were also included as a comparison group.
Further details and models used can be found in the Appendix.


\vspace{-0.13cm}
\begin{table}[!htbp]
\centering
\scalebox{0.695}{
\begin{tabular}{c|cccccccc}
\toprule
~ & BadNets & Troj-WM & Troj-SQ & Nature & Blend & $l_2$ inv & $l_0$ inv \\ 
\hline
ACC   & \textbf{94.10} & \textbf{98.85}   & \textbf{98.76} &  \textbf{98.66} & \textbf{97.00} & \textbf{98.85} & \textbf{98.86}\\
BDR   & \textbf{90.50} & \textbf{99.99}   & \textbf{99.82} &  \textbf{99.61} & \textbf{96.30} & \textbf{99.99} & \textbf{100}  \\
ACC*   & 49.76 & 85.17   & 55.37 &  54.19 & 64.52 & 77.31 & 49.08\\
BDR*   & 1.38 & 72.19   & 12.59 &  10.24 & 30.90 & 56.46 & 0.00  \\
\bottomrule
\end{tabular}}
\scalebox{0.695}{
\begin{tabular}{c|cccccccc}
\toprule
~ & BadNets & Troj-WM & Troj-SQ & Nature & Blend & $l_2$ inv & $l_0$ inv \\ 
\hline
ACC   & \textbf{90.23} & \textbf{93.96}  & \textbf{93.93} &  \textbf{91.46} & \textbf{93.67} & \textbf{93.96} & \textbf{93.93}\\
BDR   & \textbf{92.55} & \textbf{100 }   & \textbf{99.94} &  \textbf{95.00} & \textbf{99.43} & \textbf{100}   & \textbf{99.94}  \\
ACC*   & 48.92 & 57.43   & 48.61 &  49.35 & 80.63 & 89.53 & 48.40\\
BDR*   & 17.42 & 31.51   & 16.92 &  18.15 & 69.91 & 84.65 & 16.57  \\
\bottomrule
\end{tabular}}
\scalebox{0.695}{
\begin{tabular}{c|ccccccc}
\toprule
~ & BadNets & Troj-WM & Troj-SQ & Nature & Blend & GAN-tri  \\ 
\hline
ACC   & \textbf{97.74} &\textbf{ 99.29}  & \textbf{99.29} & \textbf{ 99.29} & \textbf{99.29} & \textbf{93.96} \\
BDR   & \textbf{96.94} & \textbf{100}    & \textbf{100}   &  \textbf{100}   & \textbf{100}   & \textbf{100 }  \\
ACC*   & 53.05 & 52.55   & 57.35 &  60.29 & 62.27 & 50.27 \\
BDR*   & 72.27 & 72.40   & 82.01 &  87.90 & 91.80 & 68.30 \\
\bottomrule
\end{tabular}}
\caption{The detection efficiency and comparisons on CIFAR-10 (top), GTSRB (middle) and PubFig (bottom) (\%). *represents the comparison group using the image domain data.}
\label{table:various-attacks}
\end{table}
\vspace{-0.671cm}

\paragraph{Results.}
A supervised detector built in the frequency domain leads to a high BDR (98.5 percent averaging), as shown in Table \ref{table:various-attacks}.
However, the image domain detector (represented by * in Table \ref{table:various-attacks}) does not work well.
We observe an increase in the BDR but a drop in the average ACC using the image data from the PubFig dataset versus the other two, indicating that the BDR improvement on the PubFig dataset causes a higher false-positive rate.

\paragraph{Remark 1.}\textit{High-frequency artifacts in existing backdoor triggers can be used to provide accurate detections.
Compared with the image domain, the frequency domain can enable more accurate rejection of backdoored data without sacrificing much of the clean samples.}








\subsection{Transferability}

This section evaluates the transferability of the frequency-based detector towards new datasets. 
The training set develops the same way as the above experiments.
We then test the detector's transferability from a CIFAR-10 model to the GTSRB dataset (Table \ref{table:gtsrb-trans}). The transferability of a model trained on GTSRB and a model trained on CIFAR-10 to the TSRD dataset (Table \ref{table:tsrd-trans}) is also tested.

\begin{table}[!htbp]
\centering
\scalebox{0.695}{
\begin{tabular}{c|cccccc}
\toprule
~ & \multicolumn{2}{c}{GTSRB} & \multicolumn{2}{c}{CIFAR-10} &  \multicolumn{2}{c}{CIFAR-10+Tune} \\
\textbf{Attack} & ACC & BDR & ACC & BDR & ACC & BDR  \\ 
\hline
BadNets    & 90.23  & 92.55  & 68.23  &  99.61  & 89.44  & 95.95  \\
Troj-WM    & 93.96  & 100    & 68.42  &  99.99  & 91.47  & 100    \\
Troj-SQ    & 93.93  & 99.94  & 68.40  &  99.96  & 91.44  & 99.95  \\
Nature     & 91.46  & 95.00  & 67.79  &  98.75  & 94.03  & 97.08  \\
Blend      & 93.67  & 99.43  & 66.51  &  96.18  & 64.49  & 45.67  \\ 
$l_2$ inv  & 93.96  & 100    & 68.40  &  99.95  & 91.45  & 99.97  \\
$l_0$ inv  & 93.93  & 99.94  & 68.41  &  99.98  & 91.46  & 99.99  \\   
\bottomrule
\end{tabular}}
\caption{The transferability using the detector trained on different datasets tested on GTSRB (\%).}
\label{table:gtsrb-trans}
\end{table}

Table \ref{table:gtsrb-trans}'s column headers indicate the training set used to train the specific detector. For the last column (CIFAR-10+Tune), we first train using the CIFAR-10 dataset, then fine-tune with a 200-sized dataset (half clean, half randomly perturbed originating from the 100 clean samples from the GTSRB test set) of the same distribution as GTSRB. 
In real life, as the defender is on the user's side, they will have access to the inference data, and a fine-tuning of the model using 100 clean samples is reasonable and practical. 
Note that the samples we use to fine-tune the models are not utilized in the test set for all experiments.

\begin{table}[!htbp]
\centering
\scalebox{0.695}{
\begin{tabular}{c|cccccccc}
\toprule
~ & \multicolumn{2}{c}{GTSRB} & \multicolumn{2}{c}{GTSRB+Tune} &  \multicolumn{2}{c}{CIFAR-10} &  \multicolumn{2}{c}{CIFAR-10+Tune} \\
\textbf{Attack} & ACC & BDR & ACC & BDR & ACC & BDR & ACC & BDR  \\ 
\hline
BadNets    & 57.99  & 86.83  & 77.01  &  87.10  & 61.17  & 98.01  & 82.53  & 89.83  \\
Troj-WM    & 64.57  & 100    & 83.46  &  100    & 62.16  & 100    & 87.10  & 98.97  \\
Troj-SQ    & 64.57  & 100    & 83.46  &  100    & 62.16  & 99.95  & 87.58  & 99.93  \\
Nature     & 60.09  & 91.03  & 83.11  &  99.29  & 59.30  & 94.28  & 79.61  & 83.98  \\
Blend      & 59.04  & 88.94  & 82.92  &  98.92  & 55.37  & 86.41  & 83.62  & 92.01  \\ 
\bottomrule
\end{tabular}}
\caption{The transferability on the TSRD dataset (\%).}
\label{table:tsrd-trans}
\end{table}

When comparing the original GTSRB detector and the CIFAR-10 detector on GTSRB, we see a significant drop in ACC resulting from the variance between the two datasets' data distribution. However, by fine-tuning the detector using the 200-sized dataset, one can achieve a higher ACC without sacrificing too much in the BDR. 
The detection efficiency is close to or even surpasses the detector's results with the original GTSRB training set on some attacks. 
The Blend attack is a particular case here, as the fine-tuned results worsen. We propose the main reason behind this is that the two datasets are of significant variance in distributions.
This side effect over the detection deficiency against Blend is recovered in the following experiments using pairs of training and testing sets with closer distributions.

Table \ref{table:tsrd-trans} presents the results on evaluating the detector's transferability from the GTSRB and CIFAR-10 datasets onto the TSRD dataset.
Due to the limited size of the TSRD dataset, we cannot achieve satisfying accuracy using the target model we present in the Appendix; therefore, the TSRD dataset is only for testing.
The raw detector results are similar to the experiment evaluating the CIFAR-10 model over GTSRB test data. After fine-tuning with 100 TSRD clean samples (dataset of size 200), both detectors can achieve satisfying detection results with acceptable ACC on the TSRD dataset. 
Of note, after fine-tuning, both detectors' performances against the Blend attack over the TSRD dataset are better than the results from the previous experiment's over the GTSRB dataset.
We believe this is because of closer similarities in distribution between the datasets than between CIFAR-10 and GTSRB.

\begin{table}[!htbp]
\centering
\scalebox{0.695}{
\begin{tabular}{c|cccc}
\toprule
~ & \multicolumn{2}{c}{Combined} & \multicolumn{2}{c}{Combined+Tune} \\
\textbf{Attack} & ACC & BDR & ACC & BDR  \\ 
\hline
BadNets    & 64.28  & 89.88  & 80.28  &  89.80  \\
Troj-WM    & 69.34  & 100    & 85.28  &  99.80  \\
Troj-SQ    & 69.34  & 100    & 85.36  &  99.95  \\
Nature     & 64.67  & 90.66  & 83.29  &  95.82  \\
Blend      & 64.18  & 89.68  & 84.61  &  98.45  \\ 
\bottomrule
\end{tabular}}
\caption{The transferability with extended training set, tested using the TSRD dataset (\%).}
\label{table:trans-size}
\end{table}

We also notice the CIFAR-10 detector achieves higher accuracy than the GTSRB detector on the TSRD dataset for most cases, even though CIFAR-10 and TSRD have disparent sample categories. 
Given that the two detectors are all trained with the same number of epochs and settings, we deduce that the transferability is related to the training set's size.
This assumption is confirmed in the following experiment when evaluating the transferability using a combined training set of CIFAR-10 and GTSRB.
As shown in Table \ref{table:trans-size}, when using a combined dataset, we can see an improvement in the average detection efficiency over the TSRD dataset.




\paragraph{Remark 2.}\textit{Since the high-frequency artifacts of existing triggers are universal across different datasets, transfer learning can be adopted in the task of detecting backdoored samples in the frequency domain. Even if the defender does not have access to the original training set, they can still effectively detect attacks and achieve satisfying results in the frequency domain by adopting large public clean datasets to conduct transfer learning.
}


\section{Creating Smooth Triggers}

\subsection{Problem Defination}
Given existing attacks' high-frequency artifacts, this section aims to create triggers invisible in high-frequency but stay efficient as backdoor triggers.
We summarize generating a smooth trigger as a bilevel optimization problem:
\begin{align}
    & \min_{\delta } \Sigma_{i} L(x_{i}+\delta,y_{tar};\theta_{p})+\lambda\Omega(\delta;g),\label{equ:upper1}\\
    &s.t.\ \ \underbrace{x_i+\delta}_{i=1,\ldots,N} \in[0,1]^{n},\label{equ:outer1}\\
    &\ \ \ \ \ \ \ \ \theta_{p}=\text{argmin}_{\theta} \left ( \Sigma_{i}L(x_i,y_i;\theta)+\Sigma_{j}L(x_{j}+\delta,y_{tar};\theta) \right )
    \label{equ:lower1}
\end{align}
We adopt $\Omega (\cdot;g)$ from SmoothFool \cite{dabouei2020smoothfool} to measure the input sample's roughness given a preset low-pass filter in the image domain $g$. $\lambda$ is the Lagrangian coefficient that controls the trade-off between smoothness and perturbation scale. Equation (\ref{equ:upper1}) is the optimization problem that tries to minimize both the loss of the poisoned data given a trained poisoned model and the roughness of the trigger itself. Equation (\ref{equ:outer1}) ensures the poisoned samples falls within the rational range from $[0,1]$. Equation (\ref{equ:lower1}) is the optimization problem to train a poisoned model where $\theta_p$ is the poisoned model, and $\theta$ is an initialized target model, where $j$ is the index of the samples selected as poison data.

\subsection{Methodology}
There are two ways to achieve the constraint of smoothness with the low-pass filter. 
One way is to conduct the search iteratively and output the results when it meets the constraint.
However, we find this methodology is ineffective in our case as optimization along the gradient of DNNs causes local impulses in the triggers that easily exceed the constraint. 
Therefore, we adopt a strategy by updating the smooth trigger with the perturbation that remains after the low-pass filter for each iteration, thus meeting the constraint. The remaining parts of the perturbation from the filter can be interpreted as $r=\delta*g$. Here, $r$ is the result of the perturbation after convolving with the low-pass filter, $g$, in the image domain. 
Taking Equation (\ref{equ:outer1}) into account and the fact that the triggers are of small values after passing through $g$, we adopt a min-max scaler, $M$, as a normalization process to remap the poison data onto the rational range of an image, $[0,1]$. 
Instead of using the rigid value clipping done in other works, we argue that normalization can better keep the relative scale between each pixel of the smooth trigger and better maintain functionality as a backdoor trigger. 
Consequently, we can rewrite the optimization as:
\begin{align}
    & \min_{r} \Sigma_{i} L(x^{poi}_i,y_{tar};\theta_{poi}),\\
    &s.t.\ \ r=\delta*g,\\
    &\ \ \ \ \ \ \ \ \underbrace{x^{poi}_{i}=M(x_i+\lambda r)}_{i=1,\ldots,N},\\
    &\ \ \ \ \ \ \ \ \theta_{poi}=\text{argmin}_{\theta}
    \left (\Sigma_{i}L(x_i,y_i;\theta)+\Sigma_{j} L(x^{poi}_{j},y_{tar};\theta)\right )
\end{align}
This bilevel optimization function's objective is to find a smooth pattern $r$ within the range of the low-pass filter $g$ that can be successfully adopted as a backdoor trigger. 
As stated in our paper's scope, the classifier $\theta$ is a DNN, thus making the optimization problem non-convex \cite{moosavi2017universal}. 
Thus, we propose Algorithm \ref{algo:AlgoS} to approximate a solution to this problem: we heuristically search for a smooth pattern that leads clean samples to the target label. 

\SetKwInput{KwParam}{Parameters}
\begin{algorithm}[]
\algsetup{linenosize=\tiny}
\small
    \caption{Generating a Smooth Trigger}
    \label{algo:AlgoS}
    \SetNoFillComment
    \KwIn{Data Points: $X\in R^{N\times H\times W\times C}$;\\ 
    \quad \quad \quad Pre-trained Classifier: $\theta$;\\
    \quad \quad \quad Desired Fooling Rate: $\gamma$;
    }
    \KwOut{Smooth Trigger: $r$; Dominante Label: $y_{tar}$;}
    \KwParam{Low-pass Filter $g$; Trade-off Controller: $\lambda$; Number of Classes: $K$}

    \BlankLine
     \tcc{Initialization}
     $r\leftarrow 0^{H\times W\times C}$\;
     $y_{tar}\leftarrow randint(K)$\;
     $\gamma^{best} \leftarrow Err(X)$\;
      \While{$\gamma^{best}<\gamma$}{
          \For{each data point $x_i \in X$}{
            \If{$\theta(M(x_i+\lambda r))!=y_{tar}$}{
          \tcc{Computing Purturbation}
          $\delta = -\triangledown L(x_i,y_{tar};\theta)$\;
          \tcc{Low-Pass Filter}
          $r = r+\delta * g$\;
          $r = r * g$\;
          }
          }
          $X_{poi} = M(subset(X)+r)$\;
          $y_{tar} = Domi(X_{poi})$\;
          \If{$Err(X_{poi})>\gamma^{best}$}{
          \tcc{Updating the Best Result}
          $\gamma^{best} \leftarrow Err(X_{poi})$\;
          $r^{best} \leftarrow r$\;
          $y_{tar}^{best} \leftarrow y_{tar}$\;
          }
      }
      \Return $r^{best}, y_{tar}^{best}$
\end{algorithm}
\vspace{-0.4cm}

Algorithm \ref{algo:AlgoS} explains the procedure of generating a smooth trigger. $Err(\cdot)$ computes the error rate, and $Domi(\cdot)$ output the mode of the labels that are different from their original ones. The algorithm first initializes a random target label and a zero-image as the trigger. 
While the error caused by the generated trigger is below the desired fool rate $\gamma$, the algorithm will iteratively compute the perturbation according to the gradients of a pre-trained model towards the target class for each sample that is not of the target label. 
The attained perturbation then passes through a low-pass filter to remove high-frequency parts. 
The smoothed perturbation is added to the trigger to update the smooth trigger. Finally, we select out a subset from all the data points to quickly estimate the new error rate. If the estimated error rate is larger than the preset threshold, we will update the best smooth trigger pairing with the dominant label. 
Upon experiments of generating a unified perturbation aiming to cause universal misclassification \cite{moosavi2017universal}, there exist several dominant labels that perturbations tend to lead to.
We compute the dominant label as the target label and pair it with the corresponding smooth trigger to achieve a more potent backdoor attack.

\begin{figure}[t]
  \centering
  \includegraphics[width=0.95\linewidth]{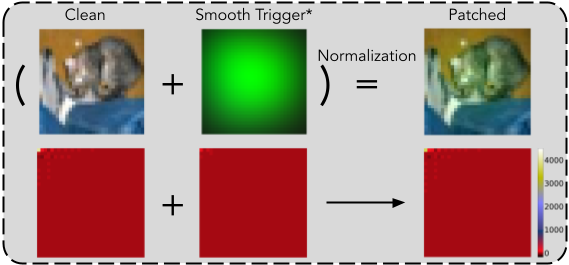}
  \caption{Visual effects over image and frequency domian of the smooth triggers. The trigger is multiplied by 5 for visualization.
  The right bottom depicts the heatmap averaged over 10000 samples patched with the smooth trigger.
  Both the trigger itself and the final images exhibit frequency spectra similar to natural images.
  }
  \label{fig:smooth}
\end{figure}

\subsection{Attack Results and Evaluations}
\label{sec:attack}

Figure \ref{fig:smooth} depicts the computed smooth trigger's visual effects using the proposed algorithm in the image domain and frequency domain. A similar figure illustrating the smooth trigger generated based on the GTSRB dataset is presented in the Appendix. 
As one can see from the frequency results, neither the trigger itself nor the final patched image contain any high-frequency components. 

We now evaluate the smooth trigger's functionality as a backdoor trigger by using it to poison the training set and conduct the entire backdoor attack pipeline. 
We adopt a small CNN trained on CIFAR-10  with an ACC of 85.50\% as the baseline model.
Then, following Algorithm \ref{algo:AlgoS}, we use the model to acquire the smooth trigger. 

The smooth attack can attain an \textit{Attack Success Rate}~(ASR) around 95\% within one epoch of training while the model's training accuracy is still below 30\%. 
This effect indicates the smooth trigger contains features that are easier to pick up by the DNN. 
We evaluate the final result when the model converges over the poison dataset with a poison ratio of 0.1\footnote{This is a standard poison rate used in other attack works \cite{li2019invisible,liu2017trojaning}.}.
The poisoned model recognizes the trigger by 97.25\% of chance and achieves an ACC on clean samples at 84.54\%, which is close to the baseline ACC.

As a comparison, we test the case of using random patches and nature images passed through the low-pass filter as naive designs of the smooth triggers. The triggers can only reach an average ASR of 75.54\%. Meanwhile, we observe that the naive-designed smooth triggers take more epochs for the model to converge. The averaging ACC over clean samples can only achieve 76.29\%, with five naive-designed smooth triggers considered. This drop in the performance over the clean samples can also impair the stealthiness of the attack.
A similar result can be witnessed on the GTSRB dataset shown in the Appendix.
Thus, we conclude that our smooth trigger maintains functionality as a backdoor trigger while leaving no high-frequency artifacts.

\vspace{-0.4cm}
\paragraph{Remark 3.}\textit{Directly using random patches passed through the low-pass filter cannot generate smooth triggers of satisfying functionality. We show that by approximately solving a bilevel problem, one can generate smooth triggers that function as backdoor triggers while achieving a satisfying stealthiness in both image and frequency domains.
}

\subsection{Impacts over Defenses}
\label{sec:impacts}

To show the importance of considering smooth triggers in defenses, we perform a small case study on \textit{Meta Neural Analysis}~(MNA) \cite{xu2019detecting}, a state-of-art defense mechanism.
When faced with a classifier poisoned with a smooth trigger, the MNA can only achieve an AUC score of 0.0776. 
However, after upgrading the MNA to consider the smooth trigger generation, the upgraded MNA can achieve an AUC score of 0.694 and a detection accuracy of 42.85\%.
This simple case study illustrates how existing defenses can be made more robust by considering smooth triggers.

Similarly, we also aim to upgrade our proposed detector with smooth triggers. 
We first try to finetune the detector with samples patched with patterns passed through the low-pass filter. 
Although the detector successfully detects samples patched with the same trigger with 95.67\% accuracy, the detector fails to generalize and cannot detect other filtered triggers nor the smooth trigger. 
We next experiment using the smooth trigger we acquired using Algorithm \ref{algo:AlgoS} to finetune the model for one epoch with 20,000 samples (half clean, half patched). 
This time, we find the model performs well on detecting the smooth trigger (82.49\% accuracy) and attains a higher detection rate of 89.37\% averaged over all unseen low-pass filtered triggers. 
With this detection rate, the detector can constrain the overall ASR of the most potent smooth trigger found using Algorithm \ref{algo:AlgoS} to 19.72\%. 
If we can use the detector to eliminate poisoned samples in the training set, we further drop the overall ASR to 18.03\%. 

\begin{figure}[t]
  \centering
  \includegraphics[width=0.99\linewidth]{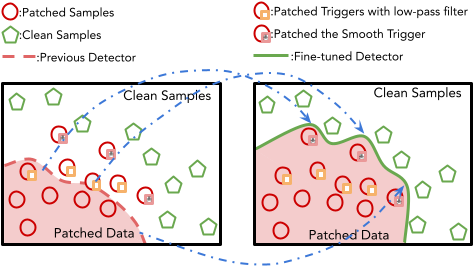}
  \caption{Fine-tuning over the smooth trigger patched samples}
  \label{fig:boundary}
\end{figure}

We design an experiment comparing the distance in the hyperplane between clean samples and samples patched with filtered triggers (including the smooth trigger and other simple designs) to better explain this generalizability.
We take the detector's last layer's weight on the benign class and compu    te the Euclidean distance between the weights and the clean samples' logits to select the ``representative'' of the clean cluster in the hyperplane. 
We then feed the poisoned samples patched with different kinds of low-pass filter processed triggers to acquire the average distance between the clean representative and the poisoned samples' clusters. 
We find that the smooth trigger patch samples have the closest distance of 4.3589 among all the filtered triggers. 
Figure \ref{fig:boundary} helps explain the generalizability acquired by fine-tuning the detector using the smooth trigger.
With a closer distance toward the clean sample center, the smooth trigger-patched samples can work as support vectors in the hyperplane to include other filtered triggers, thus achieving universal generalizability.


\vspace{-0.4cm}
\paragraph{Remark 4.}\textit{We show that defenses designed with the frequency domain considered can better mitigating the smooth triggers.
We bring attention to the development of frequency-constraint triggers, as they can be adopted in an adversarial training manner to help defenses acquire robust and generalized protection against smooth triggers.
}

\vspace{-0.2cm}

\section{Conclusion}

In this work, we filled the gap in existing works on backdoor attacks and defenses by presenting a comprehensive analysis of the overlooked frequency domain.
Unlike natural images, we found many existing attack triggers exhibit severe artifacts in the high-frequency spectrum.  
We took advantage of the artifacts and show that we can achieve an average detection rate of 98.50\% under attack-agnostic settings.
Realizing this limitation in the current trigger design, we proposed an effective way to generate triggers invisible in the high-frequency domain. We demonstrated its potency in terms of stealthiness and attack efficiency. 
Finally, we showed that existing backdoor defenses could benefit from considering frequency-invisible attacks.
We hope the remarks and solutions proposed in this paper can inspire more advanced studies on backdoor attacks in the future.


\section{Acknowledgemetns}

We would like to also thank NSF CNS-1930041 and support from Mcity.

{\small
\bibliographystyle{ieee}
\bibliography{egbib}
}

\clearpage
\appendix

\section*{Appendix}

\addcontentsline{toc}{section}{Appendices}
\renewcommand{\thesubsection}{\Alph{subsection}}

\subsection{Type-II 2D-DCT Algorithm}
The type-II 2D-DCT is given by a function $D: \mathbb{R}^{N1\times N2}\rightarrow \mathbb{R}^{N1\times N2}$ that maps an image data $\left \{ g_{x,y} \right \}$ to its frequency representation $D = \left \{ D_{k_x,k_y} \right \}$ with $D_{k_x,k_y}=$ 
{\footnotesize
\begin{align*}
\setlength{\abovedisplayskip}{2pt}
\setlength{\belowdisplayskip}{2pt}
w(k_x)w(k_y)\!\!\sum_{x=0}^{N_1-1}\!\!\sum_{y=0}^{N_2-1}\!\!g_{x,y}cos\left [ \frac{\pi}{N_1}(x+\frac{1}{2})k_x \right ]\!\! cos\left [ \frac{\pi}{N_2}(y+\frac{1}{2})k_y \right ]
\end{align*}
} 
, for $\forall k_x=0,1,...,N_1-1$ and $\forall k_y=0,1,...,N_2-1$, where $w(0)=\sqrt{\frac{1}{4N}}$ and $w(k)=\sqrt{\frac{1}{2N}}$ for $k>0$.

\subsection{Visual Examples of Different Triggers}
\begin{figure*}[!htbp]
  \centering
  \includegraphics[width=0.89\textwidth]{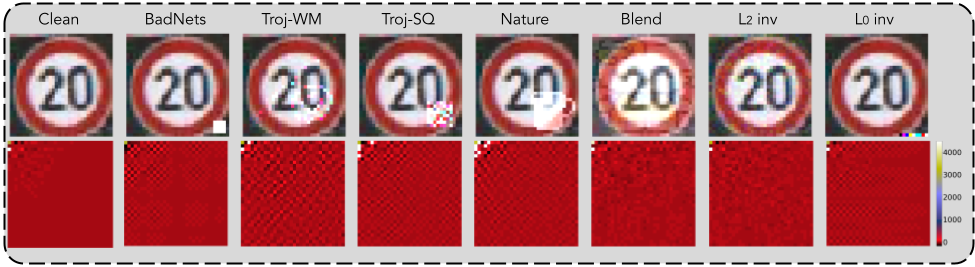}
  \caption{A pair-to-pair comparison of clean data and samples patching with different triggers on the GTSRB dataset. The frequency results are averaged over 10000 randomly selected samples from the test set.}
  \label{fig:gtsrb-frequency}
\end{figure*}

\begin{figure*}[!htbp]
  \centering
  \includegraphics[width=0.65\textwidth]{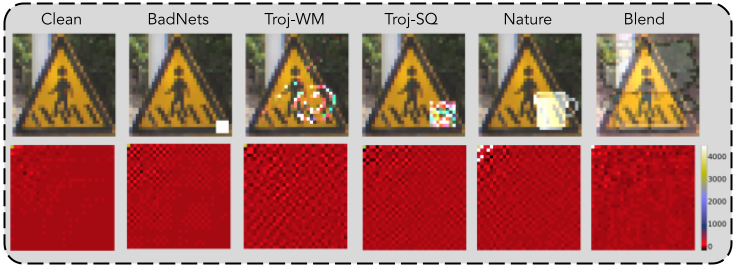}
  \caption{A pair-to-pair comparison of clean data and samples patching with different triggers on the TSRD database. The frequency results are averaged over all 4170 samples.}
  \label{fig:tsrd-frequency}
\end{figure*}

\begin{figure*}[!htbp]
  \centering
  \includegraphics[width=0.89\textwidth]{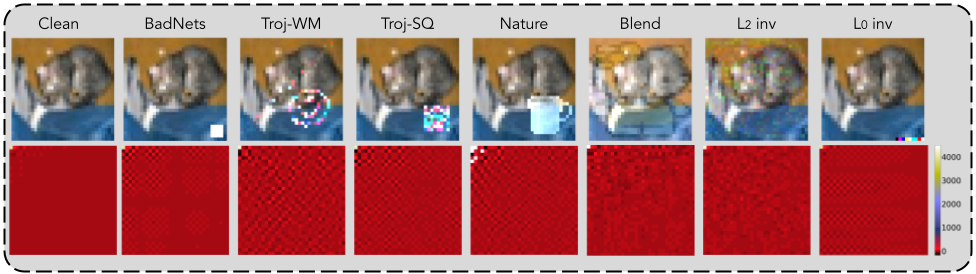}
  \caption{A pair-to-pair comparison of clean data and samples patching with different triggers on the Cifar10 dataset. The frequency results are averaged over 10000 randomly selected samples from the test set.}
  \label{fig:cifar-frequency}
\end{figure*}

\begin{figure*}[!htbp]
  \centering
  \includegraphics[width=0.79\textwidth]{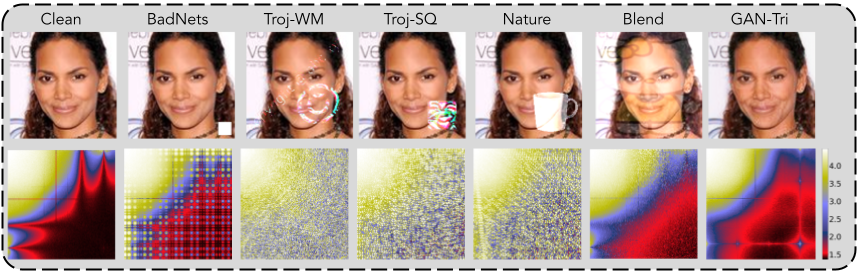}
  \caption{A pair-to-pair comparison of clean data and samples poisoned with different backdoor attacks on the PubFig dataset. The frequency results are averaged over 1000 randomly selected samples from the test set and clipped with the range of (1.5,4.5) for visualization.}
  \label{fig:pubfig-frequency}
\end{figure*}

We provide the pair-to-pair comparisons of samples patched with different triggers' visual effects in the image and frequency domain. Figure \ref{fig:gtsrb-frequency}, \ref{fig:tsrd-frequency} illustrate the comparison of the attack cases over the GTSRB and the TSRD dataset. We can see severe high-frequency artifacts similar to the CIFAR-10 dataset results presented in Section \ref{freq_exam}. We also provide the pair-to-pair extended comparison of both the image and frequency domain visual effects on the evaluated CIFAR-10 and PubFig dataset in Figure \ref{fig:cifar-frequency}, \ref{fig:pubfig-frequency}. Those results over different datasets and different triggers are provided here to further support the existence of persistent high-frequency artifacts of previous backdoor attacks in Section \ref{freq_exam}.

\subsection{Visual Examples of the Random Puturbation used in Developing the Detector}
\begin{figure}[!htbp]
  \centering
  \includegraphics[width=0.69\linewidth]{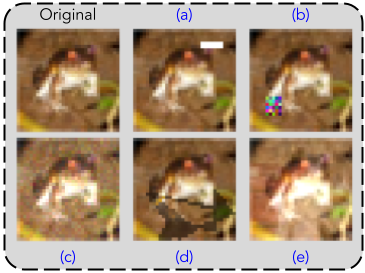}
  \caption{Visual examples of the random purturbations adopted in developing the detector. The upper left sample is a clean example, (a)-(e) are the perturbed results using different approaches.}
  \label{fig:example_purturb}
\end{figure}

Figure \ref{fig:example_purturb} presents the visual examples of the random perturbation results mentioned in Section \ref{detection_method}. Figure \ref{fig:example_purturb} (a) is the example of patching a white rectangle of random size onto a random location of the image; Figure \ref{fig:example_purturb} (b) is the result of patching a rectangle of random size and random value to a random place. Those two random perturbations simulate patching localized triggers as mentioned and analyzed in Section \ref{frequency_analysis}. Figure \ref{fig:example_purturb} (c) is the visual result of adding random Gaussian noise; the result of drawing a random shadow of random shape is depicted in Figure \ref{fig:example_purturb} (d); finally, \ref{fig:example_purturb} (e) shows the visual result of random blend.

Note that the random perturbations used in Section \ref{detection_method} as illustrated here are of different shape and values from the tested triggers. We only use those random perturbations to simulate the resulting high-frequency artifacts using the two major patching methods, as analyzed in Section \ref{frequency_analysis}.

\subsection{Linear Separability \& Input Space}

\begin{figure}[!htbp]
  \centering
  \includegraphics[width=0.99\linewidth]{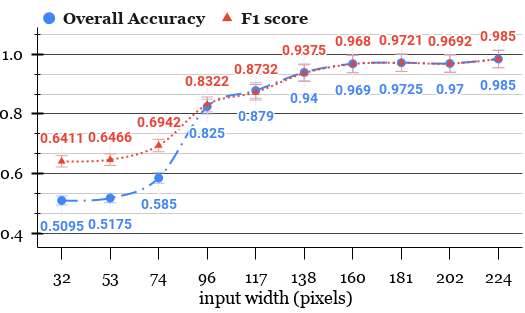}
  \caption{Detection Efficiency Using the Linear Model vs.  Input Width}
  \label{fig:width_linear}
\end{figure}

As mentioned in Section \ref{detection_method}, we look into the relationship between the input space's size and linear models' efficiency. We test the F1-score and the linear models' overall accuracy on detecting triggered samples using different-input-spaced PubFig datasets. We test ten different values ranging from 32 to 224. The relationship between the input width and the detection efficiency is depicted in Figure \ref{fig:width_linear}. We can tell from the results that a larger-input-space samples can more easily be used to conduct a linear separation of the benign samples and the triggered samples.
Meanwhile, the small-input-spaced samples are harder to be separated with linear models. Intuitively, we conduct the DCT of the whole image, thus acquiring a result of the same size as the image domain. So the larger input-spaced samples have more pixels representing the high-frequency coefficients, thus better reflecting the high-frequency artifacts when triggers are introduced. Based on the results shown in Figure \ref{fig:width_linear} and as claimed in Section \ref{detection_method}, an input space larger than 160 pixels can help linear models meet satisfying detection results.

\subsection{DNN Model Architechures and Ablation Study}

Given the different scales of difficulties to separate the DCT data in the frequency domain, we introduce a model ablation study to acquire the most simplistic DNN architecture  that satisfies the detection performance to conduct the experiments in Section \ref{detection_method}. 

\begin{table*}[t]
\centering
\scalebox{0.695}{
\begin{tabular}{c c c cccc cccc cccc}
\toprule
\multirow{2}[2]{*}{Model} & \multirow{2}[2]{*}{\#Parameters} & \multirow{2}[2]{*}{Train ACC} & \multicolumn{2}{c}{BadNets} & \multicolumn{2}{c}{Troj-WM} & \multicolumn{2}{c}{Troj-SQ} & \multicolumn{2}{c}{Nature} & \multicolumn{2}{c}{$l_2$ inv} & \multicolumn{2}{c}{$l_0$ inv}\\
 & & & ACC   & BDR   & ACC   & BDR   & ACC   & BDR   & ACC   & BDR   & ACC   & BDR   & ACC   & BDR  \\
\midrule
Linear                     & 6,146   & 83.35 &    53.85 & 28.41 & 89.64 & \textbf{100}   & 89.42 & \textbf{99.56} & 89.57 & \textbf{99.85} & 89.64 & \textbf{100}   & 64.65 & 50.00 \\
128-cell-hidden            & 393,602 & 88.23 &    54.80 & 21.89 & \textbf{93.85} &\textbf{ 99.99} & \textbf{93.44 }& \textbf{99.16} &\textbf{ 93.71} & \textbf{99.71} & \textbf{93.84} & \textbf{99.96} & 55.61 & 23.50 \\
3-layer CNN, $k_{max}=32$  & 10,214  & \textbf{95.55} &    83.64 & 72.85 & \textbf{97.21} & \textbf{99.99} & \textbf{96.94} & \textbf{99.47} & \textbf{97.09} & \textbf{99.76} &\textbf{ 97.21} & \textbf{99.99} & 70.72 & 47.03 \\
3-layer CNN, $k_{max}=64$  & 31,862  & \textbf{97.15} &    84.26 & 71.72 & \textbf{98.40} & \textbf{99.99} & \textbf{98.21} &\textbf{ 99.60} &\textbf{ 98.26} & \textbf{99.72} &\textbf{ 98.38} & \textbf{99.95} & 55.71 & 14.61 \\
3-layer CNN, $k_{max}=128$ & 109,718 & \textbf{98.36} &    86.28 & 75.44 & \textbf{98.55} & \textbf{99.99} & \textbf{98.40} & \textbf{99.68} & \textbf{98.40} & \textbf{99.67} & \textbf{98.55} & \textbf{99.99} & \textbf{97.46} & \textbf{97.80} \\
4-layer CNN, $k_{max}=128$ & 245,014 & \textbf{98.44 }&    87.63 & 78.18 & \textbf{98.52} & \textbf{99.97} & \textbf{98.36} & \textbf{99.65} & \textbf{98.39} & \textbf{99.70} & \textbf{98.53} & \textbf{99.99} & \textbf{95.25} & \textbf{93.43} \\
5-layer CNN, $k_{max}=128$ & 278,870 & \textbf{98.58} &    87.26 & 77.33 & \textbf{98.52} & \textbf{99.97} & \textbf{98.38} & \textbf{99.57} & \textbf{98.44} & \textbf{99.69} & \textbf{98.58} & \textbf{99.96} & 89.88 & 82.56 \\
6-layer CNN, $k_{max}=128$ & 292,002 & \textbf{98.64} &    \textbf{94.10} & \textbf{90.50} & \textbf{98.85} & \textbf{99.99} & \textbf{98.76} & \textbf{99.82} &\textbf{ 98.66} & \textbf{99.61} &\textbf{ 98.85} & \textbf{99.99} & \textbf{98.86} & \textbf{100}   \\
\bottomrule
\end{tabular}}
\caption{Model ablation study using the CIFAR-10 dataset. $k_{max}$ represents the maximum value of the CNN kernels. We start the analysis from the most straightforward fully-connected linear model. Hidden layers, convolutional layers, or kernel sizes are gradually added or enlarged to test out the most simplistic model that can satisfy an outstanding detection efficiency. We present the training ACC, detection ACC, and BDR for each attack (\%); the \textbf{boled} results are larger than 90\%, which we interpret as satisfying results.}
\label{tab:model_ablation}
\end{table*}

On large-input-spaced samples, namely the PubFig dataset, a linear model would already be able to achieve an outstanding detection efficiency which is introduced in Table \ref{table:various-attacks}, Section \ref{detection_results}. Thus, no further ablation study is necessary for the large-input-space. The details of the linear model we adopted to conduct the detection task over the PubFig dataset are shown in Table \ref{table:size224}. We use Adam with a learning rate of 0.01 as the optimizer for training this linear model. The binary cross-entropy is adopted as the loss function for the task of linear separation. We train the linear model with 50 epochs on the PubFig based dataset to attain the results shown in Table \ref{table:various-attacks}, Section \ref{detection_results}.

Given that the DCT results in our evaluation have the same size as the original data's input space, the DCT results over small-input-space have a weaker ability to depict high-frequency artifacts compared to larger-input-space due to the limited number of high-frequency coefficients. Thus, as shown in Table \ref{tab:model_ablation}, a similar fully connected linear model cannot meet a satisfying detection efficiency over the frequency domain using the same framework we proposed in this paper. We then conduct a thorough model ablation study by adding hidden layers or convolutional layers with different kernel sizes to obtain a most simplistic model that meets satisfying detection results over the evaluated attacks as shown in Table \ref{tab:model_ablation}. With more complex architecture and parameters, the DNN can better detect the tested attacks. Based on the ablation study, we found that only until the model's architecture consists of 6 convolutional layers with $k_{max}=128$ can it meet a satisfying and robust detection efficiency against all evaluated attacks.

\begin{table}[!htbp]
\centering
\scalebox{0.8}{
\begin{tabular}{c}
\toprule
\hline
Input $(224\times 224 \times 3)$ \\ 
\hline
Flatten (150528) \\
\hline
Dense (2) \\
\hline
\bottomrule
\end{tabular}}
\caption{The network architecture of our simple Linear detector for large input space. We report the size of each layer.}
\label{table:size224}
\end{table}

\begin{table}[!htbp]
\centering
\scalebox{0.8}{
\begin{tabular}{c}
\toprule
\hline
Input $(32\times 32 \times 3)$ \\ 
\hline
Conv2d $3\times 3$ $(32\times 32 \times 32)$ \\
\hline
Conv2d $3\times 3$ $(32\times 32 \times 32)$ \\
\hline
Max-Pooling $2\times 2$ $(16\times 16 \times 32)$ \\
\hline
Conv2d $3\times 3$ $(16\times 16 \times 64)$ \\
\hline
Conv2d $3\times 3$ $(16\times 16 \times 64)$ \\
\hline
Max-Pooling $2\times 2$ $(8\times 8 \times 64)$ \\
\hline
Conv2d $3\times 3$ $(8\times 8 \times 128)$ \\
\hline
Conv2d $3\times 3$ $(8\times 8 \times 128)$ \\
\hline
Max-Pooling $2\times 2$ $(4\times 4 \times 128)$ \\
\hline
Flatten (2048) \\
\hline
Dense (2) \\
\hline
\bottomrule
\end{tabular}}
\caption{The network architecture of our simple CNN detector for small-input-space. We report the size of each layer.}
\label{table:size32}
\end{table}

The details of the simple 6-layer CNN detector for the small-input-space are explained in Table \ref{table:size32}. The above experiments over the small-input-space are  evaluated using this model to demonstrate the efficiency of conducting the detection of backdoor triggers in the frequency domain as elaborated in Section \ref{detection_results}. We use Adam with a learning rate of 0.05 as the optimizer to train this model. Other settings are the same as the experiment conducted in large-input-space. The model took 150 epochs over the training set created using CIFAR-10 to converge and attain the results shown in Table \ref{table:various-attacks}, Section \ref{detection_results}.

\subsection{Target Model for Evaluating the Smooth Trigger}

In Section \ref{sec:attack}, we evaluate the proposed smooth attack's attack efficiency on the CIFAR-10 and GTSRB dataset. As suggested in Algorithm \ref{algo:AlgoS}, conducting the proposed attack requires a pre-trained model to generate the gradients for solving the bilevel optimization problem. We explain the details of the pre-trained model in Table \ref{table:pretrain_model}. The model was trained using Adam optimizer with a learning rate at 0.05 for 150 epochs to converge. The base-line ACC over clean samples is 85.50\% for the CIFAR-10 dataset. We also trained a base-line model on the GTSRB dataset for generating the smooth trigger over the GTSRB dataset. The GTSRB base-line model's ACC is 97.45\%.

\begin{table}[!htbp]
\centering
\scalebox{0.8}{
\begin{tabular}{c}
\toprule
\hline
Input $(32\times 32 \times 3)$ \\ 
\hline
Conv2d $3\times 3$ $(32\times 32 \times 32)$ \\
\hline
Conv2d $3\times 3$ $(32\times 32 \times 32)$ \\
\hline
Max-Pooling $2\times 2$ $(16\times 16 \times 32)$ \\
\hline
Conv2d $3\times 3$ $(16\times 16 \times 64)$ \\
\hline
Conv2d $3\times 3$ $(16\times 16 \times 64)$ \\
\hline
Max-Pooling $2\times 2$ $(8\times 8 \times 64)$ \\
\hline
Conv2d $3\times 3$ $(8\times 8 \times 128)$ \\
\hline
Conv2d $3\times 3$ $(8\times 8 \times 128)$ \\
\hline
Max-Pooling $2\times 2$ $(4\times 4 \times 128)$ \\
\hline
Flatten (2048) \\
\hline
Dense (10) \\
\hline
\bottomrule
\end{tabular}}
\caption{The target model for evaluating the smooth trigger on Cifar10 and GTSRB dataset. We report the size of each layer.}
\label{table:pretrain_model}
\end{table}

\subsection{Smooth Trigger on the GTSRB Dataset}

\begin{figure}[!htbp]
  \centering
  \includegraphics[width=0.95\linewidth]{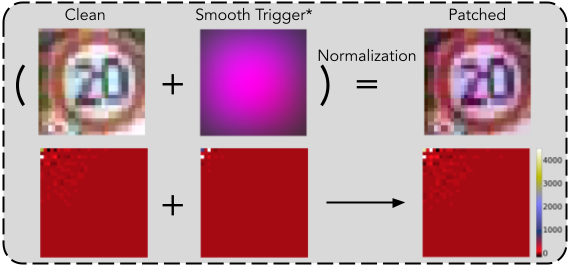}
  \caption{Visual effects over image and frequency domian of the smooth triggers. The trigger is multiplied by 5 for visualization.
  The right bottom depicts the heatmap averaged over 10000 samples patched with the smooth trigger.
  Both the trigger itself and the final images exhibit frequency spectra similar to natural images.
  }
  \label{fig:gtsrb_smooth}
\end{figure}

As mentioned in \ref{sec:attack}, we conduct the smooth attack over the GTSRB dataset following the same pipeline as well. Figure \ref{fig:gtsrb_smooth} depicts the generated smooth trigger's visual results using the GTSRB dataset in the image and frequency domain. The dominant label computed using the Algorithm \ref{algo:AlgoS} is $1$ on the GTSRB pre-trained model. Similar to the attack evaluation pipeline explained in Section \ref{sec:attack}, we conduct the backdoor attack with a poison rate of 0.1 over the target model using the GTSRB dataset. The model trained over the poisoned GTSRB dataset can maintain an ACC over clean samples at 97.42\%, which is almost the same as the base-line model. Meanwhile, the ASR is 97.86\% without defense. We observed the model could achieve an ASR greater than 90\% even with one epoch of training. Meanwhile, the detection rate of the proposed detector in Section \ref{detection_method} can only achieve a BDR at 55.31\% and an F1 score at 0.664 before considering this smooth attack. This detection efficiency can only drop the attack success rate of this GTSRB smooth trigger to 40.97\%.

By incorporating this strongest smooth trigger found using Algorithm \ref{algo:AlgoS} into the  development of the detector, we can regain a high efficient detection efficiency of a BDR at 85.53\% and an F1 score of 0.8628 using the fine-tuning pipeline proposed in Section \ref{sec:impacts}. This fine-tuning does not affect much over the other attack trigger's detection efficiency due to the limited scale as discussed in Section \ref{sec:impacts}. Using this upgraded detector on the poisoned model, we can finally constrain the ASR from 97.86\% to 13.27\% by only adopting the detector to reject samples with triggers during the inference. In the case where we apply the detector to the training phase , we can further drop the ASR to 13.03\%. 

Overall, we observe very similar results to the attack conducted over the CIFAR-10 dataset. The GTSRB datasets' results further support the remarks mentioned in the paper and emphasize the importance of the frequency domain to the development of backdoor attacks and defenses.

\end{document}